\documentclass[preprint, 12pt, hidelinks]{elsarticle}

\usepackage{amssymb}
\usepackage[utf8]{inputenc}
\usepackage{hyperref}
\usepackage{textcomp}
\usepackage{soul}
\usepackage{dirtytalk}
\usepackage{booktabs}
\usepackage[dvipsnames]{xcolor}

\usepackage[acronym]{glossaries}

\newacronym{rcc}{RCC}{Renal Cell Carcinoma}
\newacronym{prcc}{pRCC}{Papillary Renal Cell Carcinoma}
\newacronym{ccrcc}{ccRCC}{Clear Cell Renal Cell Carcinoma}
\newacronym{chrcc}{chRCC}{Chromophobe Renal Cell Carcinoma}
\newacronym{cdrcc}{cdRCC}{Collecting Duct Renal Cell Carcinoma}
\newacronym{trcc}{tRCC}{Tubulocystic Renal Cell Carcinoma}
\newacronym{onco}{ONCO}{Oncocytoma}
\newacronym{wsi}{WSI}{Whole histological Slide Image}
\newacronym{ssl}{SSL}{Self-Supervised Learning}
\newacronym{semi-sl}{semi-SL}{Semi-supervised learning}
\newacronym{cnn}{CNN}{Convolutional Neural Networks}
\newacronym{he}{H\&E}{Hematoxylin and Eosin}
\newacronym{roi}{ROI}{Region Of Interest}
\newacronym{mil}{MIL}{Multiple Instance Learning}
\newacronym{ai}{AI}{Artificial intelligence}
\newacronym{vit}{VIT}{Vision Transformer}

\journal{Computer Methods and Programs in Biomedicine}

\begin{document}

\begin{frontmatter}



\title{Renal Cell Carcinoma subtyping: learning from multi-resolution localization}




\author[inst1]{Mohamad Mohamad}
\author[inst2]{Francesco Ponzio}
\author[inst2]{Santa Di Cataldo}
\author[inst3]{Damien Ambrosetti}
\author[inst1]{Xavier Descombes}

\affiliation[inst1]{organization={Université Côte d'Azur, INRIA, CNRS},
city={Sophia Antipolis},
country={France}}
            
\affiliation[inst2]{organization={Politecnico di Torino},
            city={Torino},
            country={Italy}}

\affiliation[inst3]{organization={Department of Pathology, CHU Nice, Université Côte d'Azur},
            city={Nice},
            country={France}}

\begin{abstract}
\textbf{Background and Objective}\\
\gls{rcc} is often diagnosed at advanced stages, limiting treatment options. Since prognosis depends on tumour subtype, accurate and efficient classification is essential. Artificial intelligence tools can assist diagnosis, yet their dependence on large annotated datasets hinders broader adoption. This study investigates a \gls{ssl} framework that exploits the multi-resolution structure of \glspl{wsi} to reduce annotation requirements while maintaining reliable diagnostic performance.\\
\textbf{Methods}\\
We developed a \gls{ssl} model inspired by the pathologist’s multi-scale reasoning, integrating information across magnification levels. Robustness and generalization were evaluated through an external validation on a public \gls{rcc} benchmark and one internal validation using cohorts from the same institution but collected in different periods, with distinct scanners and laboratory workflows.
\\
\textbf{Results and Conclusions}\\
The proposed SSL approach demonstrated stable classification performance across all validation settings, reducing dependence on manual labels and improving robustness under heterogeneous acquisition conditions. These findings support its potential as a generalizable and annotation-efficient strategy for \gls{rcc} subtype classification.
\end{abstract}


\end{frontmatter}


\section{Introduction}
\label{sec:intro}
\gls{rcc} corresponds to a group of malignant tumors and the most widespread type of kidney cancer, accounting for 90\% of the overall entities. It is the 7th most common histological type in the west, with ever-increasing diffusion~\cite{RCC}. This tumor is typically asymptomatic at the early stages for most patients~\cite{RCC, truong2011immunohistochemical_RCC_med1}, which typically delays the diagnosis to a late stage of the disease, when the chances of success of the therapy are significantly lower. Because of this, the mortality-to-incidence ratio is especially unfavorable.

In accordance with WHO classification (WHO 2022), \gls{rcc} can be categorized into multiple histological subtypes. The three main ones are \gls{ccrcc}, \gls{prcc} and \gls{chrcc}, respectively accounting for for 75\%, 10\% and 5\% of the overall cases. Some of the other sutypes include \gls{cdrcc}, \gls{trcc}, and unclassified~\cite{RCC}. Approximately 10\% of renal tumors belong to the benign entities neoplasms, being \gls{onco} the most frequent subtype with an incidence of 3–7\% among all renal epithelial neoplasms~\cite{tickoo1998discriminant_RCC_diff_diagnosis2, truong2011immunohistochemical_RCC_med1}.
These subtypes show different cytological signature as well as histological features~\cite{truong2011immunohistochemical_RCC_med1}, which ends up in significantly different prognosis.

The correct categorization of the tumor subtype is indeed of major importance, as prognosis and treatment approaches depend on it and on the disease stage. For instance, the overall 5-year survival rate significantly differs among the different histological subtypes, being 55–60\% for \gls{ccrcc}, 80–90\% for \gls{prcc} and 90\% for \gls{chrcc}. This stresses the importance of an accurate subclassification \cite{fenstermaker2020development_AI1, tabibu2019pan_RCC_AI2}. 

Existing literature emphasizes also the critical role of the differential diagnosis between chromophobe and oncocytoma, known to be arduous and prone to errors due to overlapping morphological characteristics in a relevant number of cases \cite{truong2011immunohistochemical_RCC_med1, rosenkrantz2010mri_RCC_diff_diagnosis1, tickoo1998discriminant_RCC_diff_diagnosis2}.

Currently, the gold standard to classify \gls{rcc} subtypes consists in the microscopic visual assessment of \gls{he} samples, performed by a pathologist though the microscope. These specimen consist most often of physical slides and, in some centers provided with scanner, of virtual slides: the so-called \glspl{wsi}.

The visual diagnosis of the large \glspl{wsi} is known to be both effort-requiring and time-consuming, resulting in poor alignment between pathologists in some cases. Ultimately, these aspects have a  great impact on diagnosis, prognosis and treatment of \gls{rcc} neoplasms~\cite{ponzio2023improving}.

Computerized methods represents an interesting opportunity to improve the productivity, as well as the objectivity, of the microscopy-based \gls{rcc} diagnosis~\cite{ponzio2023improving}. In this regards, recent evidences suggest that \glspl{cnn}, the most popular class of supervised deep learning algorithms, may be proficiently applied to the classification of \gls{rcc} subtypes~\cite{ponzio2023improving}. This is mainly due to the \glspl{cnn}' capability to discover unseen data structures and extract robust features representation~\cite{ponzio2023improving, ponzio2019dealing}. Nonetheless, much of their strength depend on the exploitation of large labelled datasets: \glspl{cnn} are data-hungry supervised algorithms that require a substantial amount of annotated training samples to learn effective data representations~\cite{mascolini2022exploiting, huang2023self}. 
This aspect is a strong limitation, especially in the histological field~, where annotating samples demands the expertise of a skilled pathologist\cite{ponzio2023w2wnet}. Each \glspl{wsi} must be visually scrutinized and divided into sub-regions that are homogeneous in tissue architecture, and then assigned a corresponding label. This approach, known as \say{\gls{roi}-cropping procedure}, is extremely tedious and time-consuming for the pathologist~\cite{ponzio2023w2wnet}. Moreover, it is prone to errors, which may compromise the model’s training phase due to inaccurate labelling.

In this regard, \gls{ssl} has been recently attracting considerable interest, due to its ability to describe complex patterns via robust featurization, without requiring any supervision in term of samples annotations. Nonetheless, most self-supervised techniques exploit natural-scene image properties, which are not suitable for histopathology specimen~\cite{mascolini2022exploiting, ciga2022self}.

This study investigates the automated classification of \gls{rcc} subtypes through a novel self-supervised paradigm inspired by the diagnostic reasoning of pathologists. In routine practice, pathologists navigate \glspl{wsi} across different magnification levels, continuously integrating global tissue context with fine-grained cytological details to reach a reliable diagnosis~\cite{komura2025machine}. In a similar manner, our solution integrates features learned at multiple magnification levels of \glspl{wsi} and achieves performance comparable to fully supervised approaches, while substantially reducing the need for manual annotation. This design choice is consistent with recent evidence showing that multi-scale representations can effectively capture complementary contextual and morphological information, improving model robustness and generalization in histopathology image analysis~\cite{d2022comparison, rival, bontempo2023graph}. By leveraging the inherent pyramidal structure of \glspl{wsi}, our \gls{ssl}-based framework encodes both global and fine-grained features, maintaining competitive accuracy while mitigating the dependence on exhaustive expert annotation, which is time-consuming and affected by center-specific technical variability~\cite{komura2025machine}. By exploiting the pyramidal structure of \glspl{wsi}, the \gls{ssl} model learns multi-resolution feature representations that capture both local and contextual information, effectively emulating the pathologist’s multi-scale reasoning process and remaining more resilient to variations in staining reagents, chromophore reactivity, and scanner calibration~\cite{komura2025machine}. As detailed in \sectionautorefname~\ref{sec:results}, external testing on a public \gls{rcc} benchmark and internal cross-scanner validation confirmed the model’s robustness and generalization capability, with smaller performance degradation than fully supervised counterparts under heterogeneous acquisition settings.

\section{Background}

\subsection{RCC subtyping}
\label{section:RCC}

Most of the studies on \gls{rcc} subtyping come from the exploitation of the TCGA database \cite{fenstermaker2020development_AI1, tabibu2019pan_RCC_AI2, gao2020renal_RCC_AI3}, without furnishing results at the \gls{wsi} level \cite{fenstermaker2020development_AI1, tabibu2019pan_RCC_AI2} or at the patient-level \cite{fenstermaker2020development_AI1, tabibu2019pan_RCC_AI2, gao2020renal_RCC_AI3}. Furthermore, much of the current literature considers only two \cite{cheng2020computational_RCC_ML2, MICCAI_rudan} (or three \cite{fenstermaker2020development_AI1, tabibu2019pan_RCC_AI2, gao2020renal_RCC_AI3, chen2021clinical_RCC_ML1}) RCC main malignant tumor subtypes: \gls{ccrcc} and \gls{prcc} (and \gls{chrcc}). 
This is mainly due to the fact that, being dedicated to malignant tumors, TCGA data portal excludes renal oncocytoma cases. 

A considerable part of these studies seek to evaluate the performance of canonical machine learning pipelines, based on morphological or textural hand-crafted features, to discriminate between two \cite{cheng2020computational_RCC_ML2, MICCAI_rudan} or three \cite{chen2021clinical_RCC_ML1} classes of RCC subtypes. A more recent study by Fenstermak and colleagues investigated a \gls{cnn}-based approach to classify a set of 3,486 patches extracted from \glspl{wsi}, representing three classes of interest: \gls{ccrcc}, \gls{prcc}, and \gls{chrcc}. The authors achieved a patch-level accuracy of up to 99\%, but did not report the \gls{wsi}-level or patient-level performance metrics~\cite{fenstermaker2020development_AI1}.

Although these works are significant in view of the feasibility of an automatic RCC subtyping framework, they suffer from two main limitations: i) they do not take into consideration the difficult differential diagnosis between \gls{chrcc} and \gls{onco}; ii) they rely on data annotation, requiring the \gls{roi}-cropping procedure previously defined.

Answering the first limitation, in 2021, Zhu et al.\cite{zhu2021development}  investigated the renal epithelial neoplasms subtype classification in more than three classes, including the oncocytoma one.   In their study, the authors designed a classification process trained on a proprietary dataset and then tested on TCGA and biopsy database. Nonetheless, the proposed strategy is fully supervised, requiring a massive \glspl{wsi} annotation to proficiently train the employed ResNet18 classifier. A more recent study by Ponzio et al., proposed a 4 classes RCC subtype categorization through a tree-based classification pipeline~\cite{ponzio2023improving}. The authors proposed a supervised deep learning solution which leverages pathologist’s expertise and decision-making algorithm to substantially improve the classification performance of state-of-the-art \gls{cnn} models. Although interesting, the proposed methodology was still based on massive labelled dataset, being built on top of several backbone \glspl{cnn} classifier.

\label{SSLH}

\subsection{Positioning SSL within the current digital pathology panorama}
Modern computational pathology is increasingly driven by large-scale \gls{wsi}~\cite{aggarwal2025artificial}. Contemporary \gls{wsi} scanners routinely produce gigapixel slides at multiple magnifications, enabling detailed analysis of entire specimens while imposing heavy demands on storage, memory, and model design. Because current \gls{ai} architectures cannot process a \gls{wsi} end to end at diagnostic resolution, most systems adopt a tile-and-aggregate strategy~\cite{ponzio2019dealing}. The slide is tessellated into patches, each patch is encoded into a feature vector, and these patch-level embeddings are combined to yield a slide-level prediction for tasks such as subtype classification or grading. Crucially, this patch-based pipeline also enables the supervised training of \glspl{cnn} by assigning labels to individual tiles extracted from the original \glspl{wsi}; this remains the most common approach in practice. This pipeline has become standard because it is computationally tractable, aligns with typical hardware constraints, and scales well to multi-institutional cohorts~\cite{aggarwal2025artificial}. Although \glspl{cnn} still dominate the landscape, two emerging trends are gaining momentum. The first trend focuses on reducing annotation requirements by leveraging unlabelled data. Methods such as weak supervision and self-supervised learning aim to learn from minimal annotations, improving efficiency and scalability while generalizing across diverse datasets. The second trend emphasizes aligning models with pathologist reasoning. Rather than just reducing annotation, this approach integrates domain-specific knowledge, such as multi-resolution analysis, spatial dependencies, and cross-scale reasoning, to more closely mirror the cognitive processes of pathologists. These models aim to improve both interpretability and performance by capturing how pathologists integrate information from various magnifications and tissue regions.
In the remainder of this section, we briefly discuss the principal \gls{ai} paradigms for large-scale \gls{wsi} analysis—\gls{mil} frameworks, graph and hypergraph reasoning, and foundation-style \gls{ssl} pretraining. We will focus on the evolution of these methods, moving from approaches that are farther from pathologist thinking towards those that are increasingly aligned with clinical reasoning, highlighting their respective advantages, limitations, and potential for mimicking expert diagnostic processes.

\subsubsection{MIL frameworks and graph/hypergraph reasoning in histology}
\gls{mil} treats each \gls{wsi} as a bag of patch instances with only slide-level labels~\cite{bontempo2023graph}. A patch encoder produces instance features that are then aggregated, via simple pooling (e.g., mean/max) or learned attention, into a slide-level prediction. Contemporary variants introduce instance selection and clustering-aware attention to highlight diagnostically salient tiles and suppress background, improving robustness to label noise and class imbalance~\cite{ko2025cluster}. Moving beyond independent tiles, graph-based formulations represent patches as nodes with edges encoding spatial or morphological proximity, enabling message passing to capture tissue architecture; hypergraph extensions further model higher-order relations among groups of patches. In this vein, Bontempo et~al.~\cite{bontempo2023graph} propose DAS-MIL, which inserts a multi-scale graph module before the \gls{mil} aggregator: patch embeddings (learned with contrastive self-supervision) are linked within and across magnifications, a graph network yields context-enriched instance features, and these are subsequently pooled at slide level. A cross-scale knowledge-distillation term aligns coarse- and fine-resolution cues, improving robustness to scale imbalance. On Camelyon16~\cite{bejnordi2017diagnostic}, DAS-MIL reported gains over standard \gls{mil} baselines (e.g., +2.7\% AUC, +3.7\% accuracy), underscoring the value of explicit spatial context and multi-resolution coupling in weakly supervised \gls{wsi} classification. Generally speaking, \gls{mil} scales to gigapixel slides under weak supervision, is hardware-friendly, and can offer interpretable tile-level relevance (via attention or instance selection). However, it is sensitive to bag-label noise and to the \say{at-least-one} assumption, i.e., the standard premise that a positive slide contains \say{at least one} truly positive tile while a negative slide contains none. This premise can be problematic when disease signals are diffuse (spread over many mildly abnormal tiles) or rare (easy to miss during tiling), and when global context matters, potentially degrading accuracy~\cite{campanella2019clinical}.

\subsubsection{Transformer-based slide encoders}
Transformer-based encoders treat a \gls{wsi} as a large set of patch tokens (often \gls{cnn}/\gls{vit} features) and use self-attention to model long-range dependencies across the slide. Spatial information is injected via 2D relative-position encodings, and scalability is achieved with hierarchical tokenization (e.g., cluster/region pooling), sparse or windowed attention, and memory-aware sampling to handle tens of thousands of tiles. Multi-resolution designs introduce cross-scale tokens and attention to couple coarse context with fine morphology, and training can be fully weakly supervised (slide labels, \gls{mil}-style heads) or follow \gls{ssl} pretraining with a lightweight transformer aggregator. In practice, these models capture global tissue architecture more naturally than independent-tile \gls{mil} and can yield interpretable saliency via attention weights or rollout, but they incur higher compute/memory costs, often require aggressive token reduction (risking information loss), and are sensitive to the choice of positional encoding, sampling policy, and scale alignment~\cite{atabansi2023survey}.

\subsubsection{Foundation-Style SSL for WSIs}
Foundation-style \gls{ssl} learns stain- and scale-robust representations from large corpora of unlabelled \glspl{wsi}, typically via contrastive, distillation or masked-prediction objectives on tiles sampled across magnifications. Colour and geometric augmentations encourage stain invariance and localization robustness; multi-resolution sampling aligns coarse context with fine morphology. A representative instance is RetCCL, a clustering-guided contrastive pretraining framework designed for whole-slide retrieval with slide-aware sampling; its authors release general-purpose encoders that transfer to downstream slide-level tasks via linear heads or light fine-tuning~\cite{wang2023retccl}. In practice, this paradigm reduces annotation burden, improves cross-cohort generalization, and stabilizes training under label noise relative to purely supervised patch encoders. However, the pretext itself is not pathologist-aligned, e.g., RetCCL’s retrieval-oriented instance discrimination does not explicitly encode diagnostic reasoning patterns such as multi-resolution navigation, magnification ordering, or inside–outside spatial containment. This may potentially limit both explainability and the capture of domain-relevant cues during pretraining. Future improvements may arise from pathologist-inspired pretexts that explicitly mimic expert behaviour (e.g., cross-scale containment checks, ordered magnification sequences, or context reconstruction from high-magnification crops), thereby aligning learned invariances with clinical reasoning and enhancing both performance and interpretability~\cite{ponzio2023improving, rival}.

\subsubsection{Pathologist-shaped SSL pretexts: aligning pretraining with diagnostic thinking}
\gls{ssl} replaces manual labels with pretext tasks, but generic objectives from natural images (e.g., rotation, jigsaw, colorization) often mismatch histology’s symmetries and texture homogeneity~\cite{rotationpredection,jigsaw,imagecolourisation,mascolini2022exploiting}. Recent work instead designs pretexts that mirror how pathologists navigate across magnifications. Koohbanani et al.~\cite{selfpath} propose magnification prediction, a multi-scale “jigmag” ordering task, and a stain-aware colorization variant, consistently outperforming raw \gls{ssl} baselines in low-label regimes. Boyd et al.~\cite{ganpath} use visual field expansion to reconstruct a wider context from a high-magnification crop with an adversarial autoencoder framework~\cite{advautoencoder}. Srinidhi et al.~\cite{rsp} first train a network on a pretext task that predicts the correct order of three patches taken at different resolutions. After this self-supervised pretraining, the model is fine-tuned on the target task with available labels, and finally refined with consistency learning in a teacher–student setup~\cite{consistency}. A common emerging thread is exploiting the slide’s multi-resolution nature; however, most methods do not explicitly teach the model to reason about the interconnections between patches taken from different magnifications. This represents a missed opportunity to model multi-resolution dependencies and the broader context across the slide, which could enhance both performance and interpretability. Addressing this, Ding et al.~\cite{rival} ask whether a zoomed-in patch lies inside or outside a zoomed-out region, a pretext closely aligned with routine diagnostic zooming that outperforms magnification prediction and standard transfer in lung adenocarcinoma subtyping.

Taking inspiration from this latter study, we propose a novel \gls{ssl} approach grounded in the concept of inter-level connectivity. Across our \gls{rcc} subtyping experiments, covering internal validation cohorts from the same institution (acquired in different periods and laboratories) and one public, external cohort~\cite{zhu2021dartmouth}, our approach performs better than the considered counterparts, as later detailed. In particular, when compared with fully supervised baselines, it is moderately worse on the source dataset; however, under domain shift to external target data, and fine-tuning with only a few patients, it becomes decisively better, indicating improved transferability and reduced dependence on exhaustive annotation.

 \section{Materials and methods}

\begin{table}[t]
\begin{center}
\begin{tabular*}{0.95\linewidth}{@{\extracolsep{\fill}}lcccccc}
\hline
& \multicolumn{3}{c}{\textbf{Nice-A}} & \multicolumn{3}{c}{\textbf{Nice-B}}\\
\cline{2-4}\cline{5-7}
\textbf{Subtype} & Train & Test & Total & Train & Test & Total\\
\hline
\gls{ccrcc} & 33 & 23 & 56 & 2 & 18 & 20\\
\gls{prcc}  & 15 & 7  & 22 & 2 & 18 & 20\\
\gls{chrcc} & 3  & 3  & 6  & 2 & 18 & 20\\
\gls{onco}  & 3  & 4  & 7  & 2 & 18 & 20\\
\hline
\textbf{Total} & 54 & 37 & 91 & 4 & 72 & 80\\
\hline
\end{tabular*}
\end{center}
\caption{Patient distribution across folds and RCC subtypes for the \emph{Nice-A} and \emph{Nice-B} cohorts.}
\label{tab:split-NiceA-NiceB}
\end{table}

\begin{table}[t]
\begin{center}
\begin{tabular*}{0.95\linewidth}{@{\extracolsep{\fill}}lcccccc}
\hline
& \multicolumn{3}{c}{\textbf{Dartmouth Resections}} & \multicolumn{3}{c}{\textbf{Dartmouth Biopsies}}\\
\cline{2-4}\cline{5-7}
\textbf{Subtype} & Train & Test & Total & Train & Test & Total\\
\hline
\gls{ccrcc} & 10 & 20 & 30 & 0 & 34 & 34\\
\gls{prcc}  & 10 & 20 & 30 & 0 & 21 & 21\\
\gls{chrcc} & 5  & 18 & 23 & 0 & 0  & 0\\
\gls{onco}  & 10 & 10  & 20 & 0 & 24 & 24\\
\hline
\textbf{Total} & 35 & 68 & 103 & 0 & 79 & 79\\
\hline
\end{tabular*}
\end{center}
\caption{Patient distribution across \emph{Dartmouth} cohorts. Fine-tuning on the external domain uses resections only (all available \gls{chrcc} cases for training and a balanced subset of the remaining subtypes); the entire biopsy cohort is held out for testing only.}
\label{tab:dartmouth_cohorts}
\end{table}

\subsection{Dataset details}
\label{sec:dataset}
\subsubsection{Pretext training cohort (Nice-A)}
\label{sec:dataset-NiceA}

In this database, hereafter \emph{Nice-A},  we collected tissue samples from 91 consecutive patients, who encountered nephrectomy in the Nice Hospital Urology Department, diagnosed with \gls{ccrcc} (n=56), \gls{prcc} (n=22), \gls{chrcc} (n=6) or \gls{onco} (n=7). As defined by the 2022 WHO criteria, the diagnosis was based on pathology and cytogenetic analysis. H\&E stained \gls{wsi} (scanned at $\times 40$ using a Leica SCN400
Digital Slide Scanner with 0.25 µm/pixel resolution, Leica Biosystems, Wetzlar, Germany) used for diagnosis were gathered to define a dataset consisting in a total of 201 \glspl{wsi}. From the whole \glspl{wsi} dataset, with the supervision of two skilled pathologists, \glspl{roi} depicting the five classes of interest (namely, healthy normal renal tissue, \gls{ccrcc}, \gls{prcc}, \gls{chrcc}, \gls{onco}) where extracted for the downstream supervised task, as later detailed. Each category is equally represented in the \glspl{roi} dataset. Note that these patients were split among non-overlapping training and test sub-cohorts for the training of the pretext task, as later described. 

\subsubsection{Internal validation cohort (Nice-B)}
\label{sec:dataset-NiceB}
The internal validation cohort, hereafter \emph{Nice-B}, comprises H\&E slides sourced from the same laboratory as \emph{Nice-A}, but with a different scanner (Aperio AT2 scanner, Leica Biosystems, Wetzlar, Germany), yielding a spatial resolution of approximately $0.25\,\mu\mathrm{m}$/pixel. Each case is represented by multiple slides derived from distinct formalin-fixed, paraffin-embedded tissue blocks and covers all \gls{rcc} subtypes included in this study, namely \gls{ccrcc} (n=20), \gls{prcc} (n=20), \gls{chrcc} (n=20) or \gls{onco} (n=20). For fine-tuning the pretext backbone, only two patients per class are used (see table \ref{tab:split-NiceA-NiceB}-right); all remaining patients are held out strictly for validation and are never included in training.

\subsubsection{External public validation cohort}
\label{sec:dataset-Darth}
As an external validation dataset, we used the public \emph{Dartmouth} benchmark~\cite{zhu2021dartmouth}, which provides separate cohorts of surgical resections and biopsies and includes all four RCC subtypes considered in this study (see Table~\ref{tab:dartmouth_cohorts}). For fine-tuning on the external domain, we randomly sampled from the benchmark’s training/validation partitions: all available chRCC cases (n=5) and 10 cases per each of the remaining subtypes. The entire biopsy cohort was held out exclusively for external evaluation and was not used for fine-tuning.

\subsection{SSL formulation}

\begin{figure}[t]
  \centering
      \includegraphics[width=0.6\linewidth]{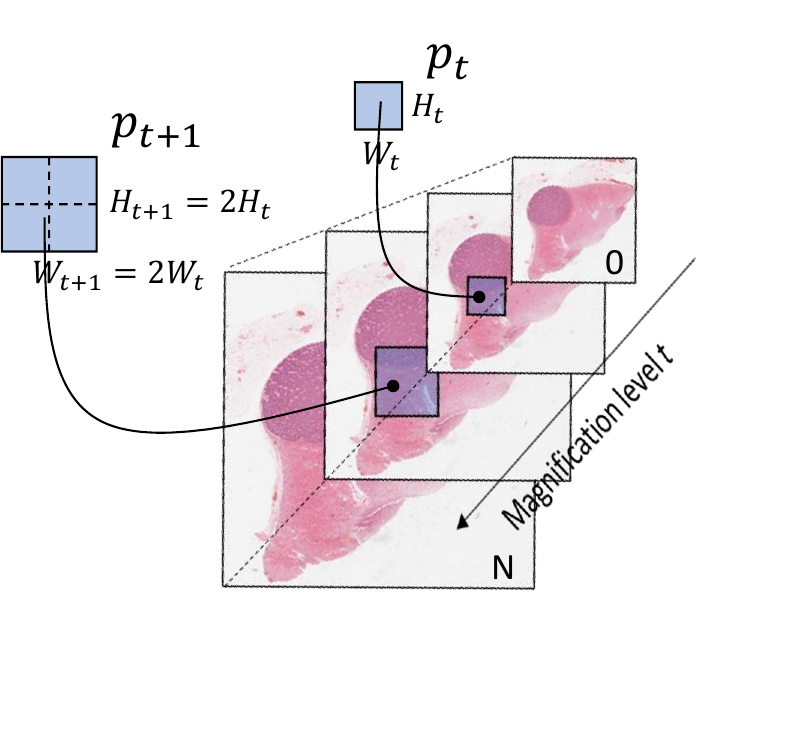}
      \caption{The typical pyramidal structure of a WSI: a collection of images that represents the same content at different magnification levels.}
      \label{fig:wsi} 
\end{figure}

\begin{figure}[t]
  \centering
      \includegraphics[width=\linewidth]{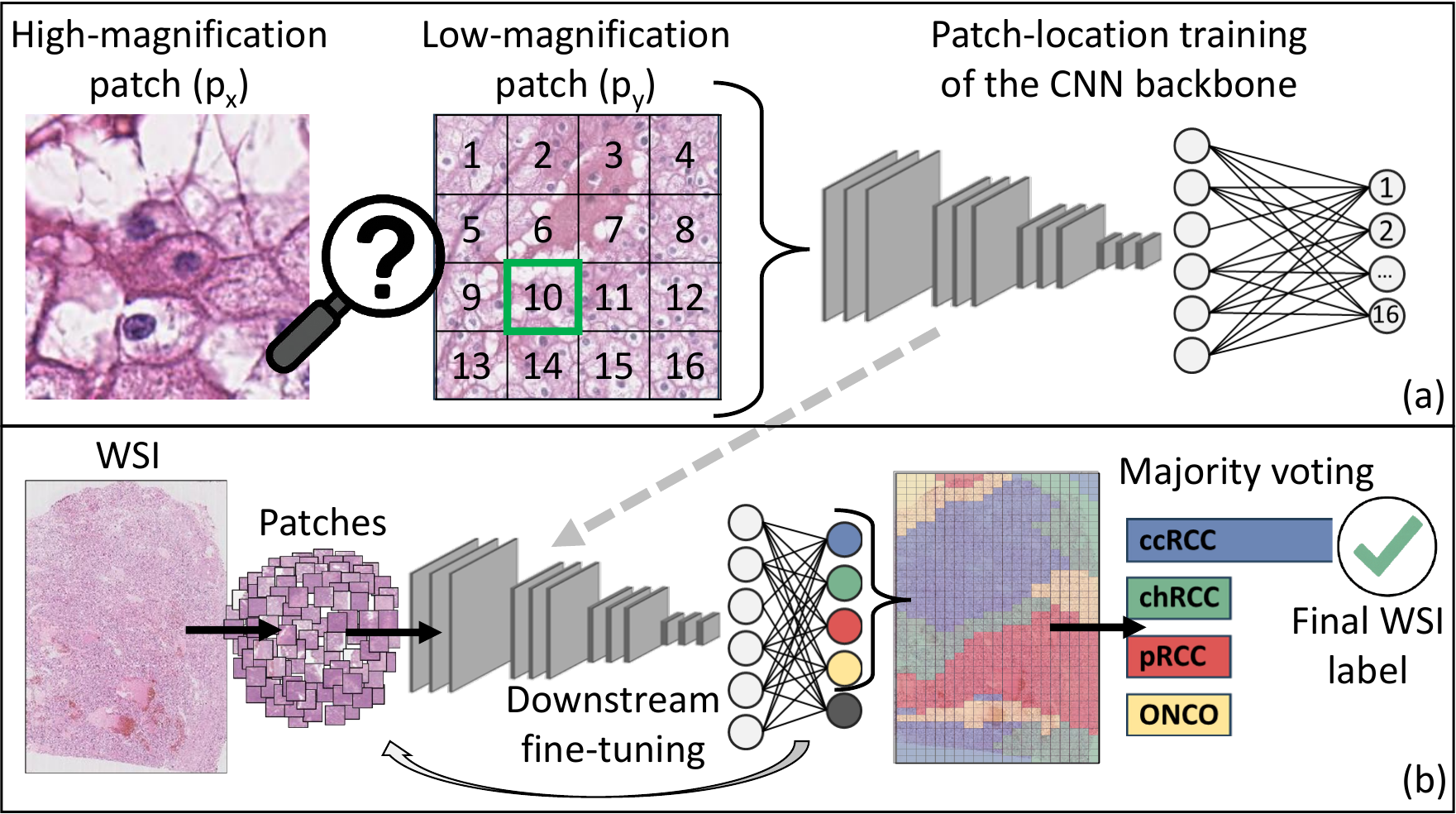}
      \caption{Overview of the proposed \gls{ssl} framework for \gls{rcc} subtype classification. (a) Pretext task for CNN pretraining: the model learns to predict the spatial location of a high-magnification patch $(p_x)$ within its corresponding low-magnification context $(p_y)$. The task exploits the multi-resolution structure of \glspl{wsi} to derive spatially consistent representations without manual annotations. (b) Downstream \gls{rcc} subtyping: the pretrained backbone is transferred and fine-tuned for classification. Patch-level predictions are aggregated through majority voting, ignoring normal tissue patches (gray neuron), to produce the final WSI-level diagnosis.}
      \label{fig:pipeline} 
\end{figure}

Suppose a WSI has $N$ magnification levels, where $0$ is the base level, featuring the lowest resolution, and $N$ is the maximum, providing the highest-resolution images of smaller areas of the tissue (see \figureautorefname~\ref{fig:wsi}). 

We can denote as $p_t$ a generic patch, taken from the WSI at level $t$, with size $(W_t, H_t)$. At a higher magnification level $t+1$, $p_t$ is represented by a higher magnification patch having size ($W_{t+1}= 2\times W_{t} \quad$, $\quad H_{t+1}= 2\times H_{t}$),
being the zoom factor between two consecutive levels $\times$2. Thus, as shown in \figureautorefname~\ref{fig:wsi}, $p_t$ can be represented as $4$ non-overlapping patches in level $t+1$, each one having the same dimensions as $p_t$. 
The formulation of our \gls{ssl} task specifically relies on such relation between patches in different WSI levels, as detailed in the following. 

Consider two resolution levels $x$ and $y \in [0 \ldots N]$, such that $y<x$, and let $p_y$ be a patch extracted from level $y$  (see \figureautorefname~\ref{fig:pipeline}(a)). We can term $C_{x|y}$ the set in level $x$ containing all the non-overlapping patches that jointly describe $p_y$. 

Lastly, let $p_x \in C_{x|y}$ be a patch belonging to level $x$, randomly extracted from $C_{x|y}$. The objective in our pre-text task is the prediction of the location of $p_x$ inside $p_y$.

We formalize our task as a classification problem by randomly sampling a patch from set $C_{x|y}$, which consists of $4^{n}$ patches, being $n=y-x$ the difference between the two magnification levels taken into account. The ground-truth label is generated using the index of $p_x$ in $C_{x|y}$. We use the cross-entropy loss function to measure the discrepancy between the predicted class probabilities and the actual class labels.
\noindent In the classification formulation, the representation of labels can pose some flexibility. Specifically, the range of the ground-truth is dependent on  $n$, which is equal to $x-y$. To ensure that such range is consistent during the training, as well as during the inference phase, we need to freeze the difference between $x$ and $y$. In other words, we are free to extract patches from different levels, but the zooming difference between the two levels must remain fixed.

\noindent To train our model on the pre-text task, we applied distinct augmentations to patches $p_x$ and $p_y$. The former is augmented both geometrically and on the color space, while the latter undergoes only color transformations (further details will follow in the next section). We found that the data augmentation phase helps to lower the over-fitting and to add robustness, avoiding simple solutions, based the tissue edge or on the white space location. Our backbone convolutional encoder, pretrained on ImageNet as later detailed, processes the augmented images using the same weights for both $p_x$ and $p_y$. Lastly, their latent vector representations are fused using a fully connected layer, following the approach of Ding et al.~\cite{rival}. 

Once proficiently trained on the pretext task, the backbone's weights are transferred on the classification level, to undergo fine-tuning on the RCC subtyping downstream task, as represented in \figureautorefname~\ref{fig:pipeline}(b).

\subsection{Implementation details}
In this section, we detail the implementation of the proposed architecture along three main pillars, as illustrated in \figureautorefname~\ref{fig:pipeline}: the pretext task, see panel~(a); the supervised \gls{rcc} classification, see panel~(b), left–middle; and the aggregation of results at the patient level, see panel~(b), right.

\subsubsection{The pretext}
We split the 91 patients between training and test set, with a diverse split ratio among the different classes to ensure proper representation of all the RCC subtypes in the test set (refer to table \ref{tab:split-NiceA-NiceB}-left). We start preparing our pre-text task dataset by fixing the zooming difference $n$ between levels $x$ and $y$ equal to two. In this way, we have 16 patches at the higher magnification corresponding to 1 at the lower. The patches $p_x$ and $p_y$, with a size of $256\times256$, are extracted randomly such that $p_x$ belongs to one of the three highest magnification levels with probabilities (0.4, 0.4, 0.2). The abundant patches available at higher magnification is the main reason for our choice. 
To properly define a supervised pretext task, we proceed in the following way with the aim of obtaining a cohort of training samples, coupled with corresponding annotations. We start by individuating $p_y$ and $C_{x|y}$ (i.e., the magnified representation of $p_y$ in the higher magnification level). Then, we randomly sampled $p_x$ from $C_{x|y}$. Lastly, its location is one-hot encoded to serve as ground truth for the cross-entropy loss.

Some filtering steps are also implemented to remove images with large white areas. The final dataset, consisting of 784,495 tiles with size $256\times256$, is further split into training and validation folds (92/8\% respectively). As mentioned before, $p_y$ is only augmented with random contrast, while $p_x$ is augmented with random flipping, rotation, contrast, and cropping. The original cropped tiles are re-scaled to $112\times112$ before augmentation, to minimize computational time. Following the findings in \cite{ponzio2023improving}, we deploy a VGG16 network as the encoder backbone, changing only the final max pooling layer to an average pooling one. As previously mentioned, both the encoder of $p_x$ and $p_y$ are trained jointly. The two latent vectors are then concatenated in a 1024 vector, fed to a fully connected layer of 256 neurons, and finally to the last layer, made up of 16 outputs (namely, 16 classes or possible locations of $p_y$ in $p_x$). Training on our pretext task required a very low learning rate of $2\times10^{-5}$, a batch of 32, and a weight decay of $10^{-5}$. We opted for an Adam optimizer. In our experiment, training reached a saddle point early on (this can be due to the small learning rate we had to start with), and defining a scheduler was non-trivial. 
Specifically, we changed the learning rate twice during training. After the first stall, we increased the learning rate to  $10^{-4}$. It must be mentioned that implementing cosine warm-up could enhance this approach, but further study is needed to determine the optimal rate of the learning rate increase. At the second stall, we increased the batch size to 64, obtaining a similar effect as decreasing the learning rate, but without suffering its long-run issues. 

\subsubsection{RCC subtype classification}
We first divided the \glspl{roi} into non-overlapping patches of size $1024\times1024$ pixels at the maximal magnification level. Patches containing predominantly white background were excluded, while the remaining tissue regions were resized to $224\times224$ pixels, which was experimentally established as the optimal input size and kept consistent across all evaluated models. Although the pretext task operated on smaller inputs ($112\times112$ pixels), this difference in patch size does not hinder the transfer of pretrained weights. Only the dimensions of the classification head need to be adjusted to match the new input resolution, while the convolutional backbone can be directly initialized with the weights learned during the self-supervised pretraining stage. During fine-tuning, we trained a five-class linear classification head comprising the four \gls{rcc} subtypes plus normal renal parenchyma; supervision derived from annotated \glspl{roi} as before detailed in \sectionautorefname~\ref{sec:dataset-NiceA}. We balance and split the resulting data into a training (85\%) and a validation subset (15\%). For the test set, we follow the same pipeline, except that we extract patches from the entirety of the \glspl{wsi}, thus without \glspl{roi}. To infer the final patient-wise label, we feed all the patches belonging to a single patient to the network and aggregate the results by a majority voting scheme, as later detailed. At aggregation time, predictions of the normal class are ignored: patches predicted as normal are discarded, and the majority vote is computed over the four \gls{rcc} subtypes only. The training of the networks, pre-trained using the self-supervised tasks, undergoes two stages. The first is an initialization for the classifier weights where we freeze the backbones and train the network for 4 epochs, the first 2 with a learning rate of $10^{-3}$, and the last 2 with a learning rate of $10^{-4}$. In the second stage, we unfreeze the backbones and deploy a cosine warmup, which allows the learning rate to reach the maximum value of $10^{-4}$ within 5\% of the total training iterations. We train the models for 120 epochs using a batch size of 2.

\subsubsection{WSI-level aggregation}
At inference time, each \gls{wsi} is divided into non-overlapping tissue patches after background removal. For each patch, the network outputs a softmax probability vector over five classes, \gls{ccrcc}, \gls{chrcc}, \gls{prcc}, \gls{onco}, and normal renal parenchyma. Before slide-level aggregation, patches predicted as normal are ignored (i.e., discarded from voting), since the downstream task is RCC subtyping. The remaining (tumor) patch labels are then used to obtain the \gls{wsi}-level classification by majority voting over the four RCC subtypes. A schematic representation of this process is provided in the right part of \figureautorefname~\ref{fig:pipeline}, panel~(b). In the rare case of a tie between two or more subtypes, we compute, for each tied class, the average softmax confidence across all patches predicted as that class and select the class with the highest average confidence as the final \gls{wsi} label.

\subsection{Counterpart models}
In this study, we compare the proposed architecture against five representative counterparts, covering the principal state-of-the-art training paradigms used for machine learning models in histopathological \gls{wsi} classification, namely:
\begin{enumerate}[i)]
    \item a canonical CNN pre-trained on the ImageNet dataset and fine-tuned on the RCC classification task;
    \item the \gls{ssl} methodology, proposed by Ding et al~\cite{rival};
    \item \emph{DAS-MIL}, the \gls{mil}-based graph neural network described in~\cite{bontempo2023graph};
    \item the contrastive self-supervised \emph{RetCCL} framework~\cite{wang2023retccl}, used to pretrain the backbone with clustering-guided contrastive learning and slide-level mosaic aggregation, followed by fine-tuning on the \gls{rcc} task;
    \item a fully supervised solution, proved to be the top performing one in this classification scenario in a recent study \cite{ponzio2023improving}.
\end{enumerate}

\subsubsection{Imagenet-based}
To define a solid transfer learning based counterpart, we refer to our previous work, where we individuated the VGG16 \gls{cnn} pre-trained on ImageNet as optimal transfer learning framework in this classification scenario~\cite{ponzio2023improving}. The model is trained with a batch size of 32 and a learning rate of  $10^{-4}$ for a total of 120 epochs. 

\subsubsection{SSL methodology}
As previously mentioned, the solution by Ding et al.~\cite{rival} consists of a model which learns a powerful data representation by understanding if a magnified patch lies inside or outside of a zoomed-out one. To put into effect such method, we started from our \gls{ssl} task dataset and we generated the data by taking a tuple of samples and switching their high-resolution patch with a probability of 0.5. If the patches were switched, both samples became negative samples, and if they were not switched, the samples were considered positive. We used the same fusion scheme and hyper-parameters they used in their experiment and trained until the performance plateau at around 50 epochs. 

\subsubsection{DAS-MIL}
We adopt the multi-scale \gls{mil} framework of Bontempo et al.~\cite{bontempo2023graph}, strictly following the authors’ public implementation and training protocol. Data preprocessing and slide-level evaluation are aligned with the original release.

\subsubsection{RetCCL}
We use the \emph{RetCCL} framework of Wang et al.~\cite{wang2023retccl}, adhering closely to the released codebase. Encoder initialization and task configuration follow the authors’ guidelines; data handling and optimization are harmonized with the other counterparts.

\subsubsection{Fully supervised solution: the ExpertDT}
As fully supervised counterpart, we select \emph{ExpertDT}, a recent classification model which leverages a combination of supervised deep learning models (specifically CNNs) and pathologist’s expertise to substantially improve the accuracy in the RCC subtyping task~\cite{ponzio2023improving}.

All counterparts use Adam with a weight decay of $10^{-5}$, and the best checkpoint is selected by validation accuracy. For embedding-based methods (ImageNet-based, Ding et al.~\cite{rival}, RetCCL~\cite{wang2023retccl} and ours), downstream classification uses a linear head. By contrast, DAS-MIL~\cite{bontempo2023graph} is evaluated in its original end-to-end multi-scale \gls{mil} configuration (graph-attention with attention-based aggregation).

\section{Results}
\label{sec:results}

\subsection{Performance on the Nice-A cohort}

A \gls{ssl} framework should ideally have two main characteristics. First, it should provide an accuracy that is not too far from the one of a fully supervised solution. At the same time, it should be robust and capable of scaling effectively when the size of the dataset used for the downstream supervised learning task is reduced. Thus, we designed two specific experiments to assess these characteristics.

\begin{figure*}[ht!]
\centering
\includegraphics[width=\columnwidth]{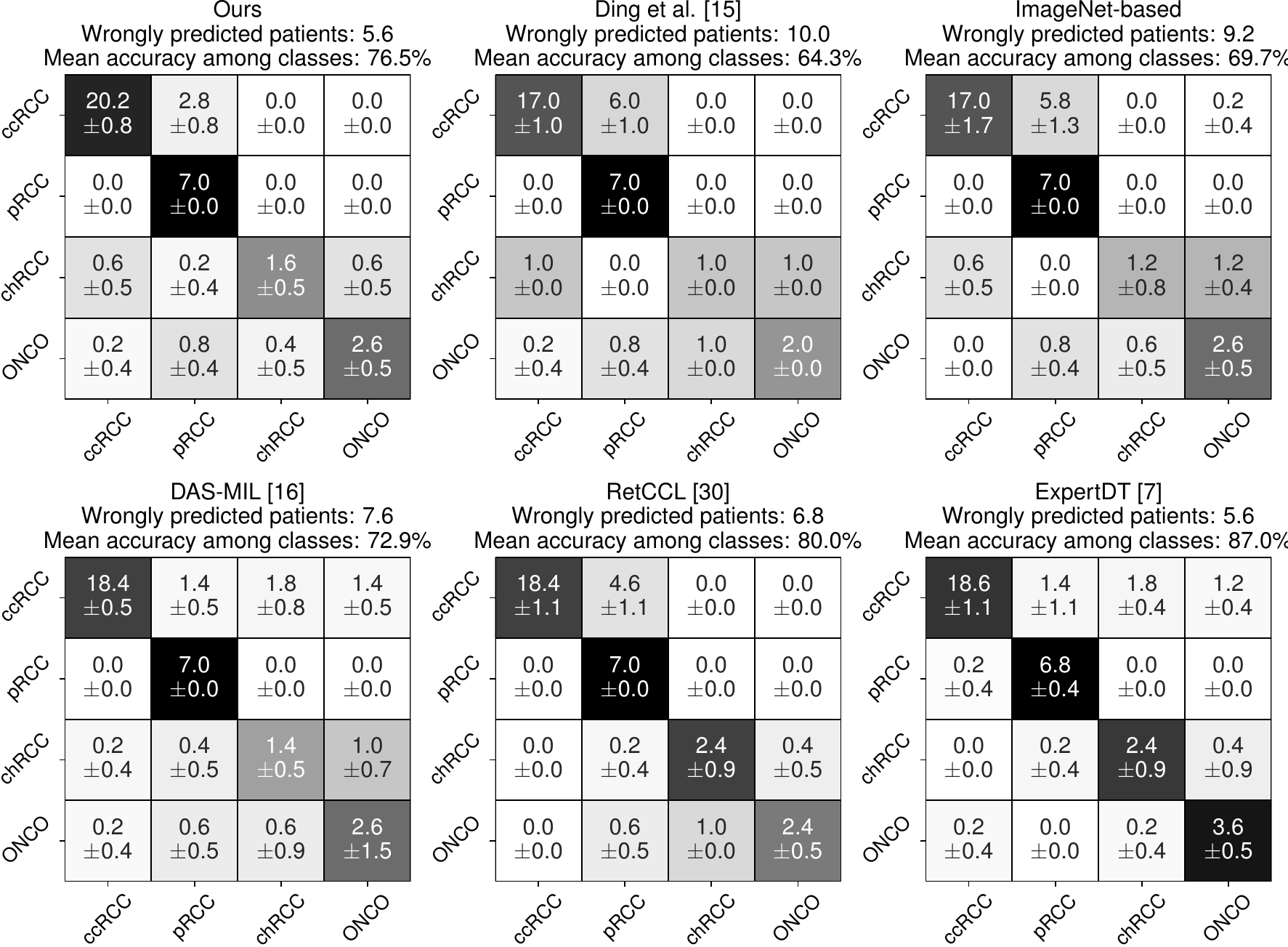}
\caption{
  \label{fig:conf-mat}
  Confusion matrices obtained on the \gls{rcc} subtyping task on the \emph{Nice-A} cohort for our method (top-left), Ding et al.~\cite{rival} (top-middle), ImageNet-based (top-right), DAS-MIL~\cite{bontempo2023graph} (bottom-left), RetCCL~\cite{wang2023retccl} (bottom-middle), and the fully supervised ExpertDT~\cite{ponzio2023improving} (bottom-right). Rows denote ground truth and columns predictions. For each approach, we report the mean accuracy among all the classes and the absolute number of misclassified patients.}
\end{figure*}

\figureautorefname~\ref{fig:conf-mat} summarizes the first experiment, displaying the confusion matrices averaged over five runs for \gls{rcc} subtyping on the \emph{Nice-A} dataset: our method (top-left) alongside all counterparts. Experiments were repeated five times according to the dataset-split protocol in Table~\ref{tab:split-NiceA-NiceB}-left. Each matrix reports, for the four target classes, the average absolute number of correctly classified patients on the main diagonal and the corresponding misclassifications off-diagonal. Repeating the study five times also enables a robustness assessment across different training runs: the variability of the patient counts per cell is quantified by the standard deviation reported in \figureautorefname~\ref{fig:conf-mat}. Notably, our approach exhibits consistently low variability together with high counts on the main diagonal (see the top-left panel of \figureautorefname~\ref{fig:conf-mat}), indicating stable performance across runs.

In terms of total misclassified patients, our approach and ExpertDT are the top performers (5.6 each), followed by RetCCL (6.8), DAS-MIL (7.6), the ImageNet-based transfer baseline (9.2), and the SSL method of Ding et al. (10).
Considering mean per-class accuracy, we rank third (76.5\%), trailing the fully supervised ExpertDT (87.0\%) and RetCCL (80.0\%), while exceeding the SSL counterpart of Ding et al. (64.3\%), the ImageNet-based baseline (69.7\%), and DAS-MIL (72.9\%). This ranking is partly driven by class imbalance: for the least represented subtypes (e.g., \gls{chrcc}, \gls{onco} with 3–4 patients in Nice-A), a single misclassified patient shifts that class accuracy by roughly 25–33 percentage points, disproportionately impacting the macro-average compared with the overall accuracy.

\begin{figure*}[ht!]
\centering
\includegraphics[width=0.65\columnwidth]{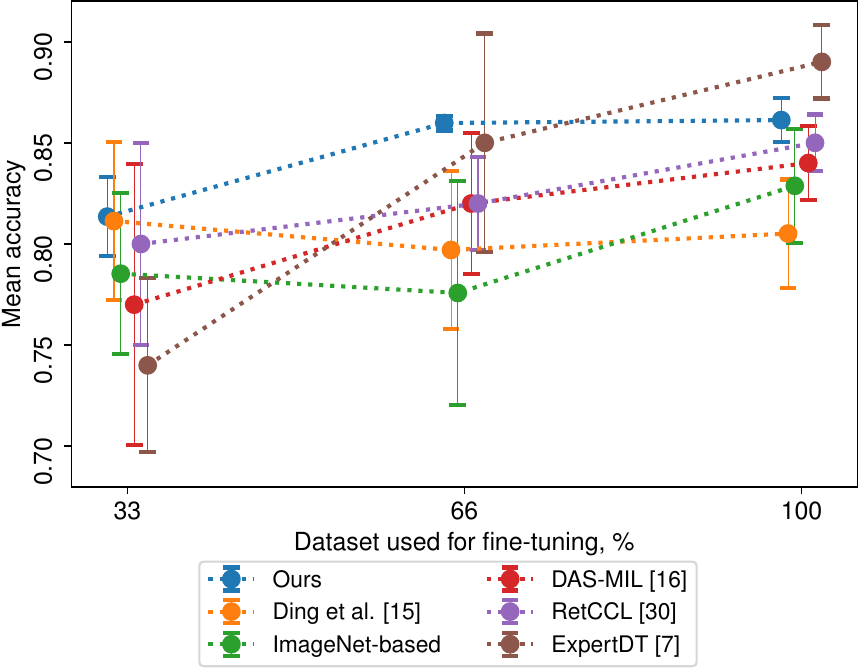}
\caption{
  \label{fig:trend}
  Mean accuracy (error bars: standard deviation over five runs) for \gls{rcc} classification as the labelled training set is progressively reduced (33\%, 66\%, 100\%). At 33\% of the data, our method is on par with the best (our and Ding et al. both at 0.81), ahead of the ImageNet-based baseline (0.79), DAS-MIL (0.77), and ExpertDT (0.74), indicating stronger data-efficiency than the fully supervised counterpart. At 66\%, our approach attains the highest accuracy (0.86), followed by ExpertDT (0.85), while RetCCL (0.82), DAS-MIL (0.82), and Ding et al. (0.80) trail behind. With the full dataset (100\%), the fully supervised ExpertDT performs best (0.89), whereas our method (0.86) remains competitive and above RetCCL (0.85), DAS-MIL (0.84), and the ImageNet-based model (0.83); Ding et al. stays flat across sizes (0.81–0.80–0.81), suggesting limited gains from additional annotation for that \gls{ssl} setup.}
\end{figure*}

With the second set of experiments, we assessed the resilience of the evaluated approaches to reductions in the size of the annotated training set used for the downstream task. \figureautorefname~\ref{fig:trend} reports the mean accuracy among classes (five runs) for \gls{rcc} subtyping as the available annotations decrease from 100\% to 66\% and 33\%. At 33\% of the data, our method reaches the top performance jointly with Ding et al. (both 0.81), ahead of the ImageNet-based baseline (0.79), RetCCL (0.80), DAS-MIL (0.77), and the fully supervised ExpertDT (0.74). At 66\%, our method attains the best result (0.86), followed by ExpertDT (0.85), while RetCCL and DAS-MIL rise to 0.82, Ding et al. slightly decreases to 0.80, and the ImageNet-based model drops to 0.78. With the full dataset (100\%), ExpertDT leads (0.89); our method plateaus at 0.86, remaining above RetCCL (0.85), DAS-MIL (0.84), the ImageNet-based baseline (0.83), and Ding et al. (0.81). Error bars in \figureautorefname~\ref{fig:trend} denote the standard deviation across five runs; overall, our method exhibits comparatively tighter intervals, especially at 66\% and 100\%, while several counterparts display wider variability at 33\%, consistent with reduced annotation. This effect is particularly pronounced for the fully supervised solution, whose dispersion increases as annotations become scarce.

\subsection{Performance on the Nice-B cohort}
\begin{figure*}[ht!]
\centering
\includegraphics[width=\columnwidth]{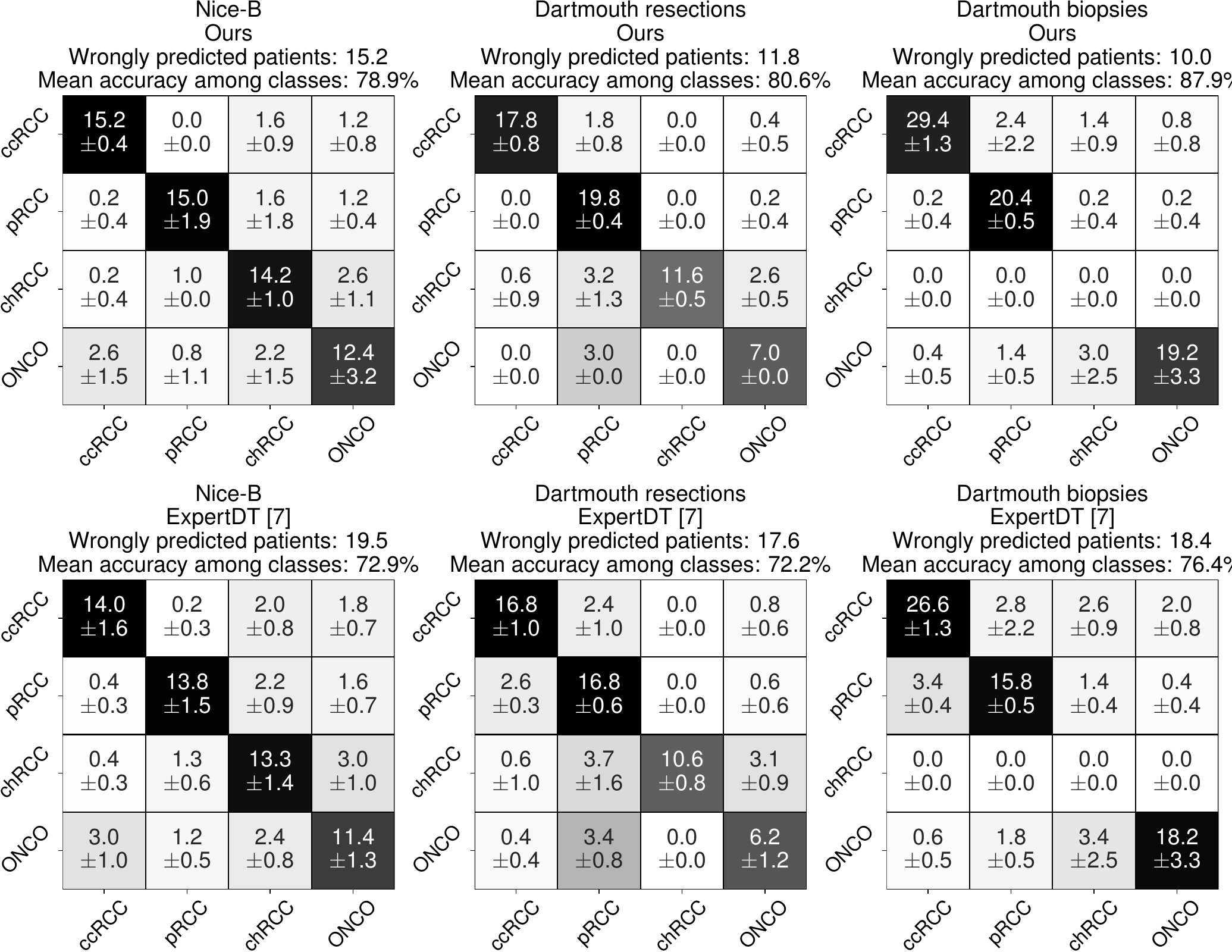}
Confusion matrices for \gls{rcc} subtyping on the \emph{Nice-B} cohort (left), the \emph{Dartmouth} surgical resections benchmark (middle), and the \emph{Dartmouth} biopsies benchmark (right). Rows denote ground truth and columns predictions. Top: our method. Bottom: ExperDT~\cite{ponzio2023improving}. For each approach, we report the mean accuracy among all the classes and the absolute number of misclassified patients.
\label{fig:all-external}
\end{figure*}

To further assess the robustness of the proposed methodology under cross-protocol generalization, we conducted an additional validation on the full \emph{Nice-B} cohort, collected at the same institution but in a different period and under a distinct acquisition protocol including a different scanner, as detailed in \sectionautorefname~\ref{sec:dataset-NiceB}. In this experiment, we compared our approach with the best-performing method on the \emph{Nice-A} cohort, namely the fully supervised ExpertDT. Both our method and the ExpertDT baseline were fine-tuned using two patients per class, with all remaining patients held out strictly for validation and never included in training. 
\figureautorefname~\ref{fig:all-external}-left shows the confusion matrices obtained by our model and ExpertDT on this dataset. Our model achieves a mean per-class accuracy of 78.9\%. The dominant error modes are \gls{ccrcc} to \gls{chrcc} (1.6), \gls{prcc} to \gls{chrcc} (1.6), \gls{chrcc} to \gls{onco} (2.6), and \gls{onco} to \gls{ccrcc} (2.6), for a mean off-diagonal mass of 15.2. Variability concentrates on \gls{onco} (diagonal std up to 3.2), while \gls{ccrcc} is comparatively stable (std 0.4).
The ExpertDT baseline attains an accuracy of 72.9\%. Its most frequent confusions are similarly distributed, yet yielding a larger off-diagonal mass (19.5). Diagonal standard deviations are of comparable magnitude on average (about 1.45 for ExpertDT vs.\ 1.63 for ours), though patterns differ by class. Overall, our method improves both mean accuracy and patients predictions and seems to be able to reduce the principal cross-protocol error modes on \emph{Nice-B}.

\subsection{Performance on the Dartmouth public benchmark}
To evaluate cross-institutional generalization, we performed an external validation on the \emph{Dartmouth} benchmark~\cite{zhu2021dartmouth}, which comprises a cohort of surgical specimens and a cohort of biopsies acquired under acquisition protocols distinct from our internal datasets (see \sectionautorefname~\ref{sec:dataset-Darth}). For comparability, both our method and the ExpertDT baseline followed the same downstream protocol for the fine-tuning. We randomly sampled from the benchmark’s training/validation partitions all available chRCC cases (n=5) and
10 cases per each of the remaining subtypes. All the resection cases in the test split were held out strictly for evaluation. On the contrary, the entire biopsy cohort was used exclusively for additional external validation, with no fine-tuning. Results are reported at patient level, using identical preprocessing, tiling, and evaluation metrics as in the internal studies, and aggregated over five runs with fixed splits.

On the \emph{Dartmouth} resection cohort (see \figureautorefname~\ref{fig:all-external}-middle), our method outperforms the ExpertDT baseline both in average accuracy and error counts. Specifically, we obtain a higher mean per-class accuracy (80.6\% vs.\ 72.2\%) and fewer misclassified patients on average (11.8 vs. 17.6), with gains distributed across all classes: ccRCC (17.8/20 correct vs.\ 16.8/20), pRCC (19.8/20 vs.\ 16.8/20), chRCC (11.6/18 vs.\ 10.6/18), and ONCO (7.0/10 vs.\ 6.2/10). In addition, diagonal standard deviations suggest more stable behaviour under cross-institutional shift, particularly for minority classes (e.g., ONCO: 0.0 vs. 1.2; pRCC: 0.4 vs. 0.6), while \gls{ccrcc} and \gls{chrcc} also exhibit slightly reduced variability with our method. Altogether, these results suggest an improved cross-protocol generalization on \emph{Dartmouth} resections.

On the biopsy subset of the \emph{Dartmouth} benchmark (see \figureautorefname~\ref{fig:all-external}-right), our method achieves a higher mean per-class accuracy and fewer errors than the fully supervised baseline. Averaged over five runs, we obtain a mean accuracy of 87.9\% with 10 wrongly predicted patients overall, whereas the baseline reaches 76.4\% with 18.4 errors. \gls{chrcc} is absent in this biopsy cohort (empty row), thus excluded from the average. Error modes concentrate on \gls{ccrcc} to \gls{prcc} and \gls{onco} to \gls{chrcc}; our model reduces these confusions compared to the baseline, leading to the lower off-diagonal mass. These results corroborate improved cross-institutional generalization on biopsies, particularly for \gls{prcc}.

\section{Discussion}
This study presents a multi-resolution \gls{ssl} framework for \gls{rcc} subtyping that exploits the pyramidal structure of \gls{wsi}. By coupling a cross-magnification pretext objective with downstream subtype classification, the approach encourages representations that are both scale-consistent and context-aware. Across all evaluations, the method is competitive on the primary cohort in macro-level metrics and delivers stronger performance at the patient level, particularly under cross-scanner and external shifts. These findings suggest that aligning features across magnifications promotes invariance to nuisance variations (e.g. scanner) while retaining morphology and tissue architecture that are relevant for subtype discrimination.

\subsection{Comparisons and robustness}
We benchmarked the framework against pathology-specific baselines within a harmonized pipeline. RetCCL provides a strong in-domain SSL encoder, while DAS-MIL is a widely used graph/MIL pipeline representative of state-of-the-art \gls{wsi} practice. In the primary cohort, RetCCL attains higher macro-accuracy, whereas our method correctly classifies more patients overall; a similar pattern is observed against DAS-MIL, which is competitive in macro-accuracy but lags at the patient level. The divergence between macro-averaged metrics and patient-level outcomes is consistent with the presence of under-represented classes (notably \gls{chrcc} and \gls{onco}): a single misclassification in a small class can disproportionately affect the macro-average, while aggregation at the patient level better reflects the clinically relevant decision unit. Under cross-scanner and external evaluations, the proposed \gls{ssl} pretraining yields smaller performance degradation than fully supervised counterparts, indicating improved robustness to domain shift.

\subsection{Annotation scaling}
We further examined label-scarce regimes via an annotation-scaling analysis. As the available annotations increase from low to moderate amounts, the proposed approach maintains favourable trends relative to pathology-specific baselines. This behaviour is consistent with the role of the pretext objective as an effective regularizer that supplies dense supervisory signals during pretraining, improving data efficiency when annotations are limited.

\subsection{Scope and limitations}
The scope of this work is deliberately centered on RCC subtyping, reflecting the characteristics of the available cohorts, limited dataset size, single-institution provenance for internal data, and class imbalance with particularly under-represented \gls{chrcc}/\gls{onco}. We do not aim to propose a pan-cancer solution at this stage. Likewise, we do not claim comprehensive coverage of downstream tasks beyond subtyping; rigorous evaluation of broader endpoints typically requires curated labels, larger multi-center cohorts, and careful control of confounding, which are beyond the present study’s design. Finally, while we included strong pathology-specific comparators (RetCCL and DAS-MIL) and a fully supervised baseline, we did not re-implement more computationally demanding hierarchical transformers, and we defer their assessment to future work.

\subsection{Future directions}
We envision three immediate extensions. First, expanding to multi-center cohorts with standardized acquisition and endpoints will enable a more systematic assessment of generalization and calibration under shift. Second, integrating the proposed multi-resolution \gls{ssl} with complementary aggregation schemes (e.g., attention-based or graph-based pooling) may further enhance reliability at the patient level, particularly under class imbalance. Third, exploring multi-task formulations that couple subtyping with additional clinically meaningful endpoints could leverage shared structure while preserving the strengths of cross-magnification pretraining.
In summary, the proposed multi-resolution \gls{ssl} framework provides a label-efficient and robust foundation for \gls{rcc} subtyping, balancing competitive macro-level performance with improved patient-level outcomes across internal, cross-scanner, and external evaluations.

\section{Acknowledgements and declarations}
The authors declare that they have no known competing financial interests or personal relationships that could have appeared to influence the work reported in this paper.
\\
Informed consent was obtained from all individual participants included in the study. All patients gave written consent for the use of tumor samples for research. The study included only the major patients. All of the samples are the property of the tissue collection of the Pathology department, which are declared annually to the French Health Ministry. The procedures followed were approved by the institutional review board of the University Hospital of Nice. This study was conducted in accordance with the Declaration of Helsinki.

\bibliographystyle{elsarticle-num} 
\bibliography{refs.bib}

\end{document}